%% file: main_arxiv.tex
\documentclass[twocolumn]{article}
\usepackage{template}

\usepackage{times}
\usepackage{cuted}
\usepackage{soul}
\usepackage{url}
\usepackage[hidelinks, colorlinks=true, linkcolor=red, citecolor=cyan]{hyperref}
\usepackage[utf8]{inputenc}
\usepackage[small]{caption}
\usepackage{graphicx}
\usepackage{amsmath}
\usepackage{amsthm}
\usepackage{booktabs}
\usepackage{algorithmic}
\usepackage[switch]{lineno}

\usepackage[dvipsnames]{xcolor}
\newcommand{\methodname}{{FLDM-VTON}\xspace}
\input{math_definition.tex}

\usepackage{framed,multirow}
\usepackage{longtable}
\newlength\savewidth\newcommand\shline{\noalign{\global\savewidth\arrayrulewidth
  \global\arrayrulewidth 1.5pt}\hline\noalign{\global\arrayrulewidth\savewidth}}

\title{FLDM-VTON: Faithful Latent Diffusion Model for  Virtual Try-on}

\author{
Chenhui Wang$^{1}$\and
Tao Chen$^{1}$\and
Zhihao Chen$^{1}$\and 
Zhizhong Huang$^{2}$\and \\
Taoran Jiang$^{3}$\and 
Qi Wang$^{3}$\and
Hongming Shan$^{1}$\thanks{Corresponding author} \\
\affiliations
$^{1}$ Institute of Science and Technology for Brain-inspired Intelligence, Fudan University\\
$^{2}$ School of Computer Science, Fudan University\\
$^{3}$ Suzhou Xiangji Technology Service Co., Ltd.\\
\emails
chenhuiwang21@m.fudan.edu.cn,\quad hmshan@fudan.edu.cn
}

\usepackage[lined,ruled,vlined]{algorithm2e}
\definecolor{mycommentcolor}{RGB}{191, 191, 191}

\SetCommentSty{mycomment}
\SetKwComment{Comment}{/* }{ */}

\begin{document}
\maketitle
\begin{strip}
    \vspace{-0.6in}
    \centering
    \begin{minipage}{\textwidth}
        \centering
    \includegraphics[width=.95\linewidth]{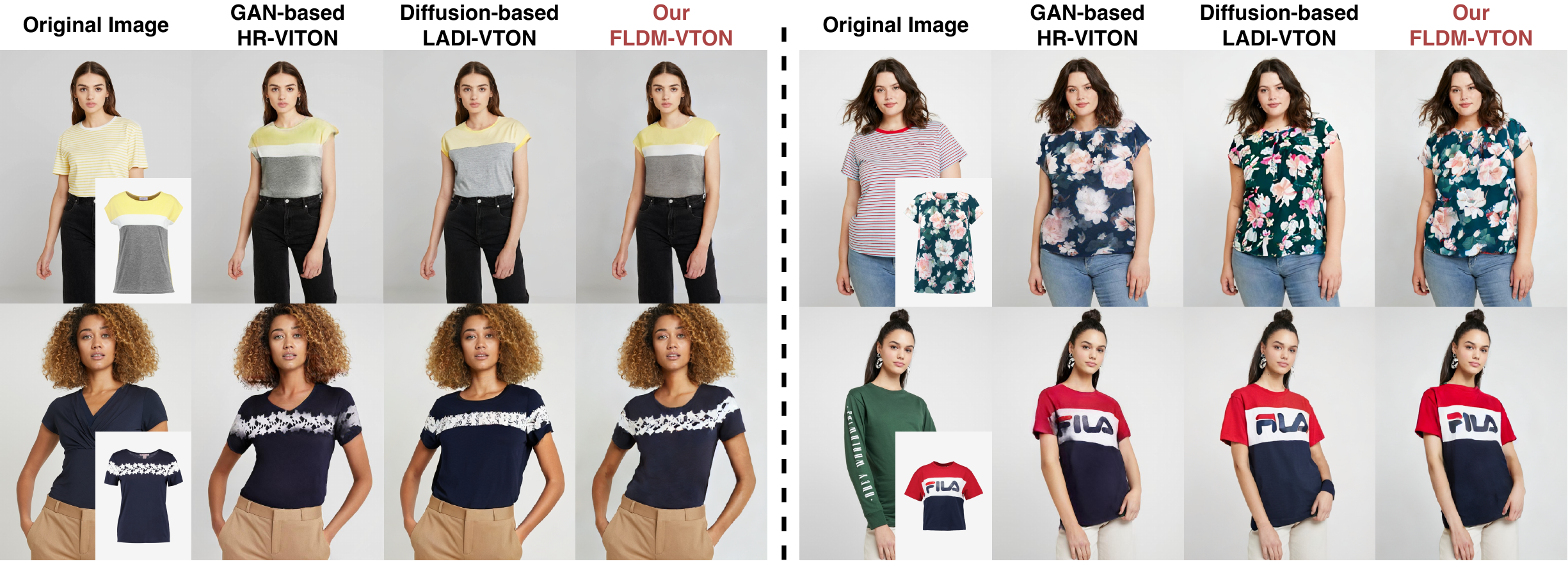}
    \vspace{-0.3cm}
    \captionof{figure}
    {
        Comparison between state-of-the-art baselines and our {\methodname} on the VITON-HD dataset.
    }
    \label{fig:first}
    \end{minipage}
\end{strip}

\input{sec/0_abstract}    
\input{sec/1_intro}
\input{sec/2_relate}
\input{sec/3_metho}

\input{sec/4_exp}
\input{sec/5_con}

\clearpage
\input{sec/6_supp}
\end{document}

%% file: math_definition.tex
\usepackage{amsmath,amsfonts}
\usepackage{amsthm}
\usepackage{amssymb,amsopn}
\usepackage{bm} 

\newcommand{\ie}{\textit{i}.\textit{e}.}
\newcommand{\eg}{\textit{e}.\textit{g}.}

\newcommand{\eat}[1]{}

\newcommand{\tabref}[1]{Table~\ref{#1}}
\newcommand{\figref}[1]{Figure~\ref{#1}}

\newcommand{\equaref}[1]{Eq.~\eqref{#1}}
\newcommand{\secref}[1]{Sec.~\ref{#1}}

\renewcommand{\paragraph}[1]{\noindent\textbf{#1}\quad}

\newcommand{\loss}[1]{\mathcal{L}_\text{#1}}
\newcommand{\lamda}[1]{\lambda_\text{#1} }   

\newcommand{\best}[1]{\textbf{#1}}
\newcommand{\suboptimal}[1]{\underline{#1}}

\newcommand{\mat}[1]{\boldsymbol{#1}} 
\newcommand{\img}[1]{\boldsymbol{#1}} 

%% file: sec/0_abstract.tex
\begin{abstract}
Despite their impressive generative performance, latent diffusion model-based virtual try-on (VTON) methods lack faithfulness to crucial details of the clothes, such as style, pattern, and text.
To alleviate these issues caused by the diffusion stochastic nature and latent supervision, we propose a novel \textbf{F}aithful \textbf{L}atent \textbf{D}iffusion \textbf{M}odel for VTON, termed \methodname. 
\methodname improves the conventional latent diffusion process in three major aspects.
First, we propose incorporating warped clothes as both the starting point and local condition, supplying the model with faithful clothes priors.
Second, we introduce a novel clothes flattening network to constrain generated try-on images, providing clothes-consistent faithful supervision.  
Third, we devise a clothes-posterior sampling for faithful inference, further enhancing the model performance over conventional clothes-agnostic Gaussian sampling.
Extensive experimental results on the benchmark VITON-HD and Dress Code datasets demonstrate that our \methodname outperforms state-of-the-art baselines and is able to generate photo-realistic try-on images with faithful clothing details.  
\end{abstract}

%% file: sec/1_intro.tex
\section{Introduction}
\label{sec:intro}
Image-based virtual try-on (VTON) aims to transfer a piece of in-shop flat clothes onto one person's body while preserving the details of both the human and the clothes, such as style, pattern, and text. In the past decade, VTON has attracted considerable attention~\cite{wang2018toward,choi2021viton,minar2020cp,lee2022high,morelli2023ladi,gou2023taming,xie2023gp}, and with the rapid advances of generative artificial intelligence~\cite{vaswani2017attention,ho2020denoising} it has the great potential to improve users' shopping experience by bridging the gap between users and online shopping.

Prior methods for VTON highly rely on generative adversarial networks (GANs)~\cite{goodfellow2014generative,huang2023adaptive,wang2024joint} to synthesize try-on images. Typically, they first use thin plate spline (TPS)-based~\cite{wang2018toward,li2021toward,fele2022c} or appearance flow-based~\cite{han2019clothflow,he2022style} algorithms to warp the flat clothes to the person's body, and then use GAN to further refine the previously generated try-on images. Nevertheless, due to the mode collapse issue~\cite{bau2019seeing}, GAN-based methods fail to synthesize photo-realistic try-on images and accurately capture intricate clothing details, often leading to flaws on the generated results; see~\figref{fig:first}.

Recently, the diffusion model has shown remarkable generative capabilities across various tasks, such as image inpainting, image editing, and even segmentation~\cite{ho2020denoising,rombach2022high,song2020denoising,chen2023berdiff}. Compared with GAN, the diffusion model offers more stable training and direct likelihood estimation. However, directly applying the diffusion model to high-resolution VTON is infeasible due to limited computational resources. 
Therefore, current diffusion-based VTON methods~\cite{morelli2023ladi,gou2023taming} are built upon the latent diffusion model (LDM)~\cite{ramesh2021zero} that performs diffusion process in a latent space. Although showing effectiveness in generating realistic try-on images, they often produce \emph{unfaithful} clothing details with respect to the original flat clothes.

We identify the \emph{stochastic nature} and  \emph{latent supervision} of LDM as the key limiting factors for the faithfulness. On one hand, the diffusion stochastic nature poses a challenge in preserving clothing details, as indicated by the initial Gaussian noise introduced at the sampling process and the added Gaussian noise at each time-step. On the other hand, the latent supervision falls short in providing image-level supervision for fine clothing details. Thus, generating highly faithful clothing details with respect to the original flat clothes using the diffusion model remains a significant challenge.

To alleviate these issues in LDM, we propose a novel \textbf{F}aithful \textbf{L}atent \textbf{D}iffusion \textbf{M}odel for VTON, termed \methodname.
To achieve faithful try-on generation, 
our \methodname improves the training of conventional latent diffusion process in two major aspects: \textbf{(i)} supplying the model with \emph{faithful clothes priors} by leveraging warped clothes as both the starting point and local condition to mitigate the initial and in-process added stochasticity, respectively, and \textbf{(ii)} providing \emph{clothes-consistent faithful supervision} through a novel clothes flattening network to bring additional image-level constraints from the original flat clothes. In addition to the training improvements, our \methodname also improves the inference process by devising a clothes-posterior sampling, further enhancing the model performance over conventional clothes-agnostic Gaussian sampling.

\paragraph{Contributions.}  Our contributions are as follows. \textbf{(i)} We propose a novel faithful latent diffusion model for VTON to address the unfaithful issue caused by the diffusion stochastic nature and latent supervision. 
\textbf{(ii)} We propose incorporating warped clothes as both the starting point and local condition, supplying the model with faithful clothes priors.
\textbf{(iii)} We introduce a novel clothes flattening network to constrain the generated try-on images, providing clothes-consistent faithful supervision. 
\textbf{(iv)} We devise a clothes-posterior sampling for faithful inference, further enhancing the model performance over conventional clothes-agnostic Gaussian sampling.
\textbf{(v)} Extensive experimental results on the VITON-HD and Dress Code datasets demonstrate that our \methodname outperforms state-of-the-art baselines and is able to generate photo-realistic try-on images with faithful clothing details.

\begin{figure*}[t]
	\centering
	\includegraphics[width=1\linewidth]{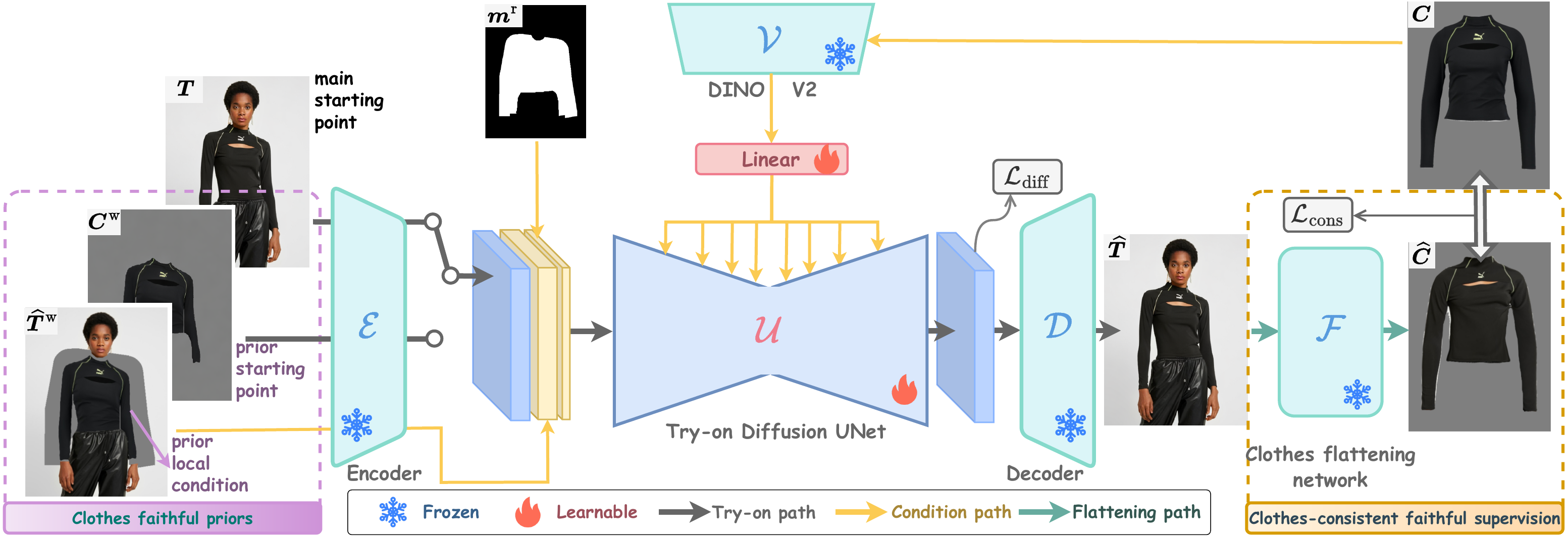}
 \caption{Overview of the proposed \methodname. Our \methodname is trained with main and prior denoising input, constrained by DINO-V2, and the proposed clothes flattening network to generate the try-on images $\widehat{\img{T}}$.
 }
\label{fig:overall} 
\end{figure*}

%% file: sec/2_relate.tex
\section{Related Work}
\label{sec:related}
\paragraph{Image-based virtual try-on.}
Numerous VTON studies have investigated generating high-resolution photo-like try-on images by transferring a in-shop flat clothes item onto one person's body, enhancing the online shopping experience. Currently, most VTON's works can be divided into two stages: pre-warping and refining. In the pre-warping stage, thin plate spline (TPS)-based~\cite{wang2018toward,li2021toward,fele2022c} or appearance flow-based~\cite{han2019clothflow,ge2021parser,he2022style} algorithms are used to warp the flat clothes to the corresponding positions on the person body. In the refinement stage, convolutional neural network (CNN) or GAN techniques are employed to further refine generated images. Although there are some diffusion-based VTON methods~\cite{morelli2023ladi,gou2023taming,Zhu2023}, they often generate unfaithful clothing details. In contrast, our \methodname prioritizes generating realistic try-on images with faithful clothing details.

\paragraph{Diffusion models.}
Inspired by non-equilibrium statistical physics, diffusion models have recently presented strong capability in image synthesis. To generate high-resolution images with limited computational resources, LDM, \eg~DALL-E and Stable Diffusion (SD) series~\cite{ramesh2021zero,rombach2022high,podell2023sdxl}, 
conducts the diffusion process in the latent space. Moreover, customized or subject-driven image generation has recently become an important research topic, including text-driven~\cite{ruiz2023dreambooth,wei2023dreamvideo} and image exemplar-based generation~\cite{yang2023paint}. DreamBooth~\cite{ruiz2023dreambooth} embeds a given subject instance in the output domain of a text-to-image diffusion model by binding the subject to a unique identifier. Paint-by-Example~\cite{yang2023paint} uses CLIP~\cite{radford2021learning} to convert the target image as an embedding for guidance. In contrast to their emphasis on preserving general subject information, our \methodname focuses on preserving more fine-grained clothing details.

%% file: sec/3_metho.tex
\section{Methodology}
\label{sec:method}

In this section, we present the try-on diffusion model with faithful clothes priors in \secref{sec:diff_main} and clothes-consistent faithful supervision in \secref{sec:takeoff}, followed by an overview of the \methodname and a clothes-posterior sampling for faithful inference in  \secref{sec:overview}.

\subsection{Try-on Diffusion with Clothes Priors}
\label{sec:diff_main}

Given a person image $\img{P}\in\mathbb{R}^{H\times W\times3}$ and a mask $\img{m}\in\{0,1\}^{H\times W}$ indicating the try-on region, one can obtain a clothes-agnostic person image $\img{P}^\text{a}$ through element-wise multiplication; $\img{P}^\text{a}$ refers to the person image with the try-on region being masked out. The goal of VTON is to transfer a piece of flat clothes $\img{C}\in\mathbb{R}^{H\times W\times3}$ onto $\img{P}^\text{a}$, yielding a photo-realistic try-on image $\widehat{\img{T}}\in\mathbb{R}^{H\times W\times 3}$ with faithful clothing details. 

Current state-of-the-art (SOTA) LDM-based VTON methods~\cite{morelli2023ladi,gou2023taming} first use TPS or flow-based warping to generate the warped clothes $\img{C}^\text{w}$ from the flat clothes $\img{C}$, and then employ an LDM for realistic refinement. 
With a pre-trained encoder $\mathcal{E}$ and decoder $\mathcal{D}$, LDM trains a  diffusion model in the latent space, involving forward  and reverse processes. 
In the forward process, Gaussian noise $\img{\epsilon}\sim\mathcal{N}(0,1)$ is added at arbitrary time-step $t$ to the resultant latent feature $\mat{z}_0=\mathcal{E}(\img{T})\in\mathbb{R}^{h\times w\times c}$, where $\img{T}$ is the ground-truth try-on image. In the reverse process, a diffusion UNet is employed to estimate the added noise $\img{\epsilon}$.

Although achieving realistic results,  conventional LDM-based VTON methods lack faithfulness to original clothing details. We identify the diffusion stochastic nature as the key limiting factor, which can be reflected in two primary aspects: \textbf{(i)} the initial Gaussian noise introduced at the sampling process and \textbf{(ii)} the added Gaussian noise at each time-step. To alleviate these, our idea is to supply the model with clothes priors, leveraging warped clothes \textbf{(i)} as the starting point to address the initial stochasticity and \textbf{(ii)} as the local condition to mitigate the in-process added stochasticity.

\figref{fig:overall} presents an overview of our \methodname. Unlike conventional forward process that takes $\mathcal{E}(\img{T})$ as the starting point, ours takes either $\mathcal{E}(\img{T})$ or warped clothes feature $\mathcal{E}(\img{C}^\text{w})$ as the starting point, in which warped clothes feature provides clothes prior at the beginning. To differentiate these two starting points, we refer to $\mat{z}_0^\text{m}=\mathcal{E}(\img{T})$ as \emph{main} starting point and $\mat{z}_0^\text{p}=\mathcal{E}(\img{C}^\text{w})$ as \emph{prior} starting point. Moreover,  we  leverage the pre-warped try-on image feature  $\mathcal{E}(\widehat{\img{T}}^\text{w})$ as prior local condition for all time-steps, where $\widehat{\img{T}}^\text{w}=\img{C}^\text{w} + \img{P}^\text{a}$. 

Next, we detail the concrete forward and reverse processes.

\paragraph{Forward process.} 
We gradually add Gaussian noise $\img{\epsilon}\sim\mathcal{N}(0,1)$ on the main and prior starting latent features, $\mat{z}_0^\text{m}$ and $\mat{z}_0^\text{p}$, with an arbitrary time-step $t$, yielding $t$-th corresponding latent features as follows:
\begin{align}
    \label{eq:ztr}
\mat{z}_t^\text{m}=\!\sqrt{\alpha_t}\img{z}_0^\text{m}+\sqrt{1\!-\!\alpha_t}\img{\epsilon},\quad\mat{z}_t^\text{p}=\!\sqrt{\alpha_t}\img{z}_0^\text{p}+\sqrt{1\!-\!\alpha_t}\img{\epsilon},
\end{align}
where $\alpha_t:= {\textstyle \prod_{s=1}^{t}}(1-\beta_s)$, and $\beta_s$ is a pre-defined variance schedule~\cite{nichol2021improved}.

\paragraph{Reverse process.} We have down-sized mask $\img{m}^\text{r}\in\{0,1\}^{h\times w}$  as the denoising condition and pre-warped try-on image feature $\mathcal{E}(\widehat{\img{T}}^\text{w})$  as the prior local condition. We concatenate the $t$-th latent feature, prior local condition, and denoising condition along the channel dimension, severing as the input to the diffusion UNet:
\begin{align}
    \img{\psi}^\text{m}_{t} = [\img{z}_t^\text{m}; \mathcal{E}(\widehat{\img{T}}^\text{w}); \img{m}_\text{r}], \quad
    \img{\psi}^\text{p}_{t} = [\img{z}_t^\text{p}; \mathcal{E}(\widehat{\img{T}}^\text{w}); \img{m}_\text{r}],
    \label{eq:gmf}
\end{align}
where $[\cdot ; \cdot]$ denotes concatenation operation.

Given one image pair, we obtain two distinct inputs: the main denoising input $\img{\psi}^\text{m}_{t}$ and the prior denoising input $\img{\psi}^\text{p}_{t}$. These inputs are individually processed through the same try-on diffusion UNet $\mathcal{U}$ to  predict the main starting latent feature ${\img{z}}^\text{m}_0$. In addition, we also encode the flat clothes $\img{C}$ through DINO-V2~\cite{oquab2023dinov2}, a currently powerful self-supervised visual encoder $\mathcal{V}$, which serves as the global controller being injected into each UNet layer via cross-attention. Therefore, the diffusion training loss function over one single sample and one  time-step $t$ is defined as follows:
\begin{align}
\label{equ:diff}
\loss{diff}=\frac{1}{2}\big(&\|\mathcal{U}(\img{\psi}^\text{m}_t, \mathcal{V}(\img{C}), t)-\img{z}^\text{m}_0\|^2_2\nonumber\\
+&\|\mathcal{U}(\img{\psi}^\text{p}_t, \mathcal{V}(\img{C}), t)-\img{z}^\text{m}_0\|^2_2\big),
\end{align}
where $\img{\psi}^\text{m}_t$ contributes to preserving the photo-realistic quality as established by existing diffusion models and  $\img{\psi}^\text{p}_t$ contributes to enhancing the faithfulness of generated clothes.

\subsection{Clothes-consistent Faithful Supervision}
\label{sec:takeoff}
Although clothes priors can help enhance the faithfulness from the input, it is still challenging to preserve the fine details such as pattern and text since the training is only supervised by the ground-truth try-on latent feature. To further improve the faithfulness to fine details, we introduce clothes-consistent faithful supervision, drawing inspiration from the fact that the clothes item you try on should be identical to the flat one once you take it off and flatten it out.  To this end, we introduce a clothes flattening network $\mathcal{F}$ that can take off clothes from the generated try-on image and flatten it out like the original flat one.

\begin{figure}[!t]
	\centering
	\includegraphics[width=1.\linewidth]{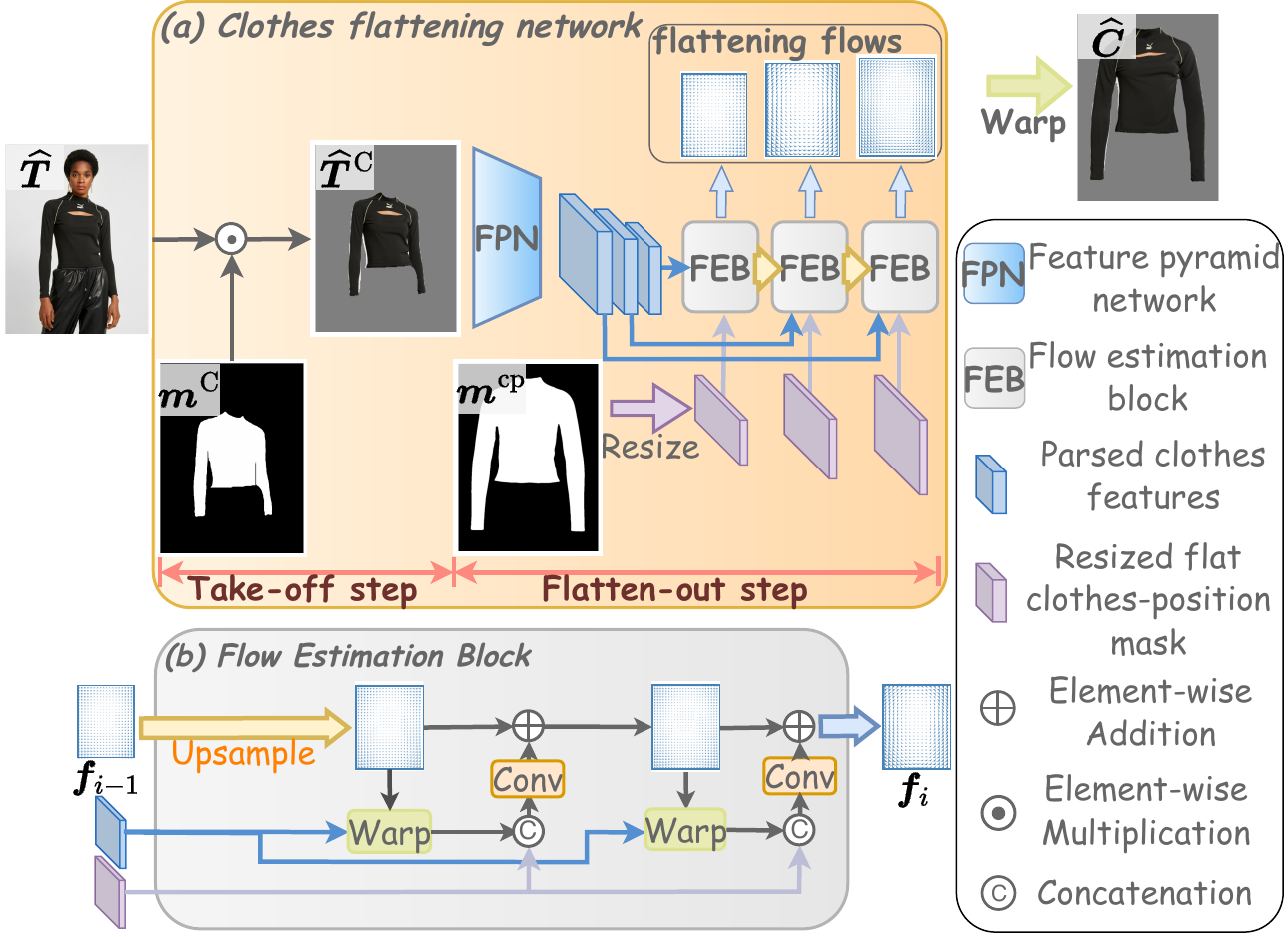}
 \caption{Illustration of the proposed clothes flattening network.}
\label{fig:takeoff} 
\end{figure}

\paragraph{Clothes flattening network.} Our clothes flattening network is a two-step method: (i) take-off step and (ii) flatten-out step. The take-off step can be easily done by masking out the generated try-on image with the pre-parsed clothes mask $\img{m}^\text{C}$. The flatten-out step is an inverse warping process, which is done by training a flattening module to predict flattening flows. More specifically, our clothes flattening network is designed with a U-shape structure, which utilizes a Feature Pyramid Network (FPN)~\cite{lin2017feature} to encode the clothes-parsed feature at multiple scales, and then employs cascaded flow estimation blocks to predict the flattening flows with down-sized flat clothes-position masks. Note that we use five different multi-scale features in our experiments; three of these are illustrated  in~\figref{fig:takeoff} for simplicity.

Following the SOTA appearance flow training strategy~\cite{ge2021parser}, we use mixed loss function to train our clothes flattening network, including $\loss{1}$ loss and perceptual loss $\loss{per}$~\cite{johnson2016perceptual} at the image level and the second-order smooth loss $\loss{sec}$ and the total-variation loss $\loss{TV}$ at the flow level, which is defined as:
\begin{align}
\label{eq:tk_flow}
\loss{flat}=\loss{1}+\lamda{per}\loss{per}+\lamda{sec}\loss{sec}+\lamda{TV}\loss{TV},
\end{align}
where $\lamda{*}$ are the hyperparameters to adjust the weights among different loss components.

\paragraph{Clothes-consistent supervision.} Once the clothes flattening network is trained, we can use the frozen clothes flattening network $\mathcal{F}$ to process the generated try-on images $\widehat{\img{T}}=\mathcal{D}(\hat{\img{z}}_0^\text{m})$, yielding estimated flat clothes $\widehat{\img{C}}$. Measuring the difference between estimated flat clothes $\widehat{\img{C}}$ and original flat clothes $\img{C}$ can further provide  clothes consistent faithful supervision for try-on diffusion training, which is defined as:
\begin{align}
\label{equ:take-off}
    \loss{cons}=\|\widehat{\img{C}}-\img{C}\|_1=\|\mathcal{F}({\widehat{\img{T}}})-\img{C}\|_1.
\end{align}

\subsection{Overview of Our \methodname}
\label{sec:overview}
\paragraph{Overview.} \figref{fig:overall} presents the overview of our \methodname. During the training phase, we handle each sample by deriving its main and prior denoising inputs, \ie~$\img{\psi}^\text{m}_{t}$ and $\img{\psi}^\text{p}_{t}$, by concatenating the $t$-th latent feature with prior local and denoising conditions. And then, our try-on diffusion UNet $\mathcal{U}$ individually processes these two types of inputs guided by the global controller DINO-V2 $\mathcal{V}$, estimating the try-on latent feature with the diffusion loss $\loss{diff}$ in~\equaref{equ:diff}. Moreover, our \methodname incorporates a clothes flattening network, providing clothes-consistent faithful supervision with the clothes-consistent loss $\loss{cons}$ in~\equaref{equ:take-off}. The overall training loss for the try-on diffusion UNet is defined as follows:
\begin{align}
    \loss{Try-on}=\loss{diff}+\lamda{cons}\loss{cons},
\end{align}
where $\lamda{cons}$ is a trade-off hyperparameter. 
Note that only the try-on diffusion UNet and the linear layer followed by the DINO V2 are trained.

\paragraph{Faithful inference.} During the inference phase, conventional diffusion models initiate inference by a clothes-agnostic noise sampled from a standard Gaussian distribution. However, this introduces significant initial stochasticity, adversely affecting the faithfulness of generated clothing details. To address this issue, we devise a clothes-posterior sampling to further enhance the model performance.

With the introduced clothes prior for model training, our \methodname can initiate inference from a posterior Gaussian noise that is specifically conditioned by the warped clothes feature $\mathcal{E}(\img{C}^\text{w})$. Specifically, the clothes-posterior noise is the $T$-th prior latent feature: $\mat{z}_T^\text{p}=\sqrt{\alpha_T}\mathcal{E}(\img{C}^\text{w})+\sqrt{1-\alpha_T}\img{\epsilon}$, corresponding to the warped clothes feature $\mathcal{E}(\img{C}^\text{w})$ after $T$ diffusion forward time-steps using \equaref{eq:ztr}. By doing this, the initial stochasticity of the sampling process is significantly reduced, thereby ensuring the faithfulness of the generated clothing details.

\begin{figure*}[!ht]
    \centering
    \includegraphics[width=1 \linewidth]{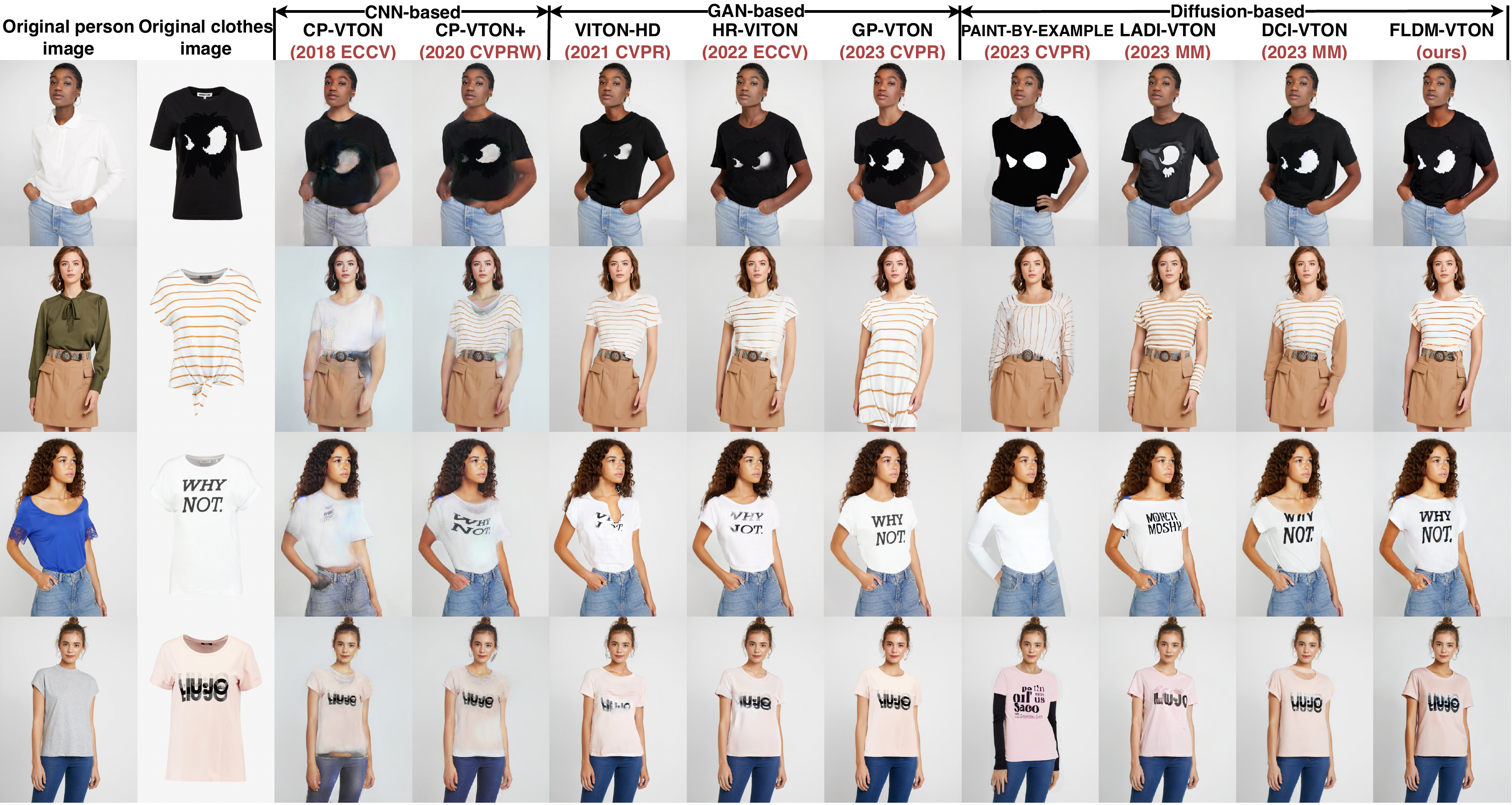}
    \caption{Qualitative results of different methods and ours on the VITON-HD dataset. Best viewed when zoomed in.}
    \label{fig:comp_vton}
\end{figure*}

%% file: sec/4_exp.tex
\section{Experiments}
\subsection{Experimental Setup}
\paragraph{Datasets.} We conduct experiments on two popular high-resolution VTON benchmarks: the VITON-HD dataset~\cite{choi2021viton} and Dress Code dataset~\cite{morelli2022dress}. Both datasets contain high-resolution paired images of size $512\times 384$:
in-shop flat clothes and their corresponding persons wearing the clothes. The VITON-HD dataset contains $13,679$ image pairs for upper-body clothes. The Dress Code dataset includes $15,366$ image pairs for upper-body clothes, $8,951$ image pairs for lower-body clothes, and $29,478$ image pairs for dresses. We follow the official guidelines to divide the data into training and testing sets~\cite{choi2021viton,morelli2022dress}.

\paragraph{Implementation details.} We use  SOTA appearance flow-based warping method~\cite{ge2021parser,gou2023taming} to generate a warped clothes image aligning the flat clothes with the person.
We adopt  Adam optimizer to optimize all networks with a mini-batch size of $8$ and a learning rate of $2.0\times10^{-5}$ on 4 NVIDIA V100 GPUs. 
In addition, we employ the encoder and decoder of  SD KL-regularized auto-encoder, with a down-sampling factor of $d =8$ and a latent channel number of $c=4$, as our encoder $\mathcal{E}$ and decoder $\mathcal{D}$, respectively.
We set $T=1,000$ for
latent diffusion training as suggested by SD~\cite{lee2022high}, and use the DPM solver~\cite{lu2022dpm} with $50$ sampling steps for inference. Besides, we use FreeU~\cite{si2023freeu} to reweight the contributions of the backbone and skip connection features with different scaling factors for inference, enhancing the denoising capability of the LDM and reducing low-frequency information; please refer to Appendix~\ref{app:imple} for more implementation details, including visualizations and detailed pipelines of $\img{P}^\text{a}$ and $\widehat{\img{T}}^\text{w}$, a detailed warping procedure, and hyperparameter setting.

\begin{table}[!t]
\centering
\resizebox{\linewidth}{!}{
\begin{tabular}{lcc|ccccc}
\shline
\multirow{2}{*}{\textbf{Methods}} &&&\multicolumn{2}{c}{\textbf{Paired}} &&\multicolumn{2}{c}{\textbf{Unpaired}} \\
&&&\textbf{LPIPS}$\downarrow$&\textbf{SSIM}$\uparrow$ &&\textbf{FID}$\downarrow$ &\textbf{KID}$\downarrow$ \\
\cline{1-3}\cline{4-5}\cline{7-8}
{CP-VTON}& \multirow{2}{*}{\textbf{\large{I}}}&&0.160&0.831&&31.34&2.37\\
{CP-VTON+}&&&0.131&0.847&&22.79&1.55\\
\cline{1-3}
{VITON-HD}& \multirow{3}{*}{\textbf{\large{II}}}&&0.116&0.862&&12.12&0.32\\
{HR-VITON}&&&0.104&0.878&&11.27&0.27\\ 
{GP-VTON}&&&\suboptimal{0.081}&\suboptimal{0.884}&&\,\,9.19&\best{0.09}\\
\cline{1-3}
{PAINT-BY-EXAMPLE} &\multirow{4}{*}{\textbf{\large{III}}}&&0.143&0.803&&11.94&0.39\\
{LADI-VTON}&&&0.096&0.863&&\,\,9.47&0.19\\
{DCI-VTON}&&&\suboptimal{0.081}&0.880&&\,\,\best{8.76}&\suboptimal{0.11}\\
{\methodname} (\textbf{ours})&&&\best{0.080}&\best{0.886}&&\,\,\suboptimal{8.81}&0.13\\
\shline
\end{tabular}
}
\caption{Comparison results on the VITON-HD dataset. \textbf{I}, \textbf{II}, and \textbf{III} represent CNN-based, GAN-based, and diffusion-based methods, respectively. The best and second best  results are highlighted in \best{bold} and \suboptimal{underlined}, respectively.}
\label{tab:comp_vton}
\end{table}

\paragraph{Evaluation metrics.} We evaluate performance under paired and unpaired test settings, where ground-truth try-on images are available for the paired setting while not for the unpaired setting. 
We adopt Learned Perceptual Image Patch Similarity (LPIPS)~\cite{zhang2018unreasonable} and Structural Similarity (SSIM)~\cite{wang2004image} for paired setting while using Fr\'echet Inception Distance (FID)~\cite{heusel2017gans} and Kernel Inception Distance (KID)~\cite{binkowski2018demystifying} for unpaired setting. 
Note that we present all qualitative results under the unpaired setting.

\begin{figure*}[!ht]
    \centering
    \includegraphics[width=1 \linewidth]{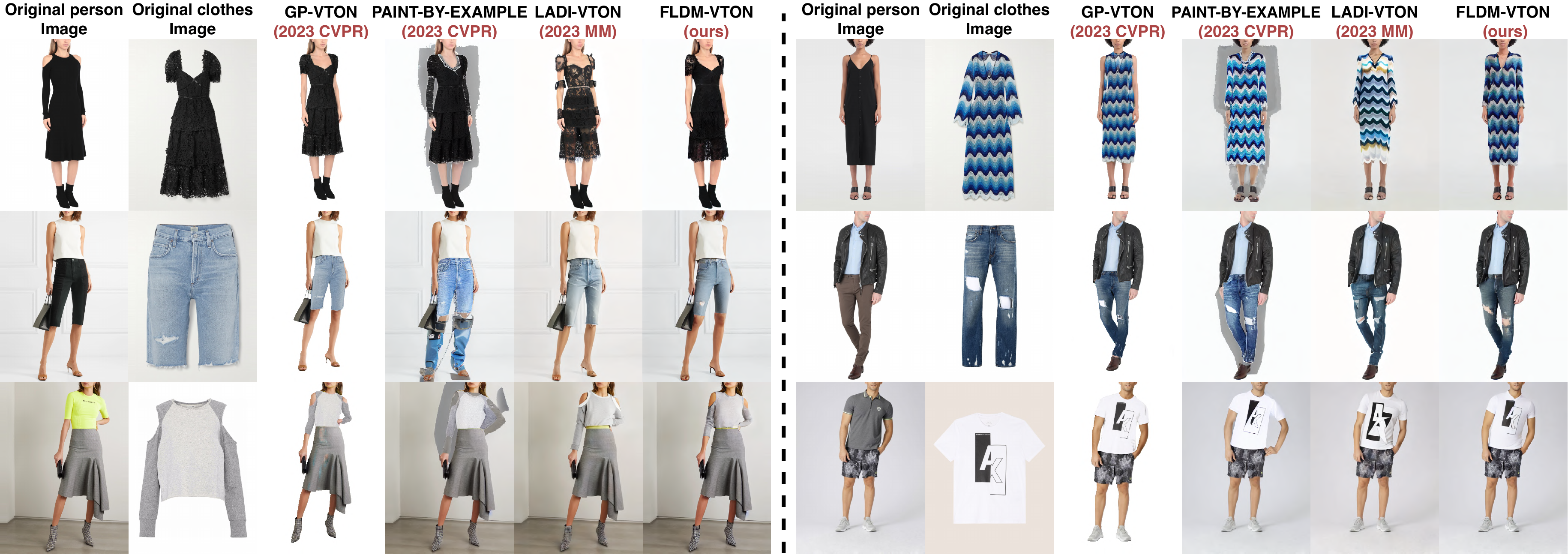}
    \caption{Qualitative results of different methods and ours on the Dress Code dataset. Best viewed when zoomed in.}
    \label{fig:comp_dress}
\end{figure*}

\begin{table*}[!ht]
\centering
\resizebox{\linewidth}{!}{
\begin{tabular}{l|ccccccccccccccccc}
\shline
\multirow{3}{*}{\textbf{Methods}} &\multicolumn{5}{c}{\textbf{Upper}} &&\multicolumn{5}{c}{\textbf{Lower}}&&\multicolumn{5}{c}{\textbf{Dresses}} \\
\cline{2-6}\cline{8-12}\cline{14-18}
&\multicolumn{2}{c}{\textbf{Paired}}&&\multicolumn{2}{c}{\textbf{Unpaired}}&&\multicolumn{2}{c}{\textbf{Paired}}&&\multicolumn{2}{c}{\textbf{Unpaired}}&&\multicolumn{2}{c}{\textbf{Paired}}&&\multicolumn{2}{c}{\textbf{Unpaired}}\\
&\textbf{LPIPS}$\downarrow$&\textbf{SSIM}$\uparrow$ &&\textbf{FID}$\downarrow$ &\textbf{KID}$\downarrow$&&\textbf{LPIPS}$\downarrow$&\textbf{SSIM}$\uparrow$ &&\textbf{FID}$\downarrow$ &\textbf{KID}$\downarrow$&&\textbf{LPIPS}$\downarrow$&\textbf{SSIM}$\uparrow$ &&\textbf{FID}$\downarrow$ &\textbf{KID}$\downarrow$ \\
\cline{1-1}\cline{2-3}\cline{5-6}\cline{8-9}\cline{11-12}\cline{14-15}\cline{17-18}
{GP-VTON}&0.271&0.775&&21.58&0.99&&0.283&0.759&&23.61&1.13&&0.221&0.789&&17.73&0.70\\
{PAINT-BY-EXAMPLE}&0.165&0.858&&38.78&2.70&&0.197&0.818&&23.56&0.99&&0.346&0.700&&21.51&0.87\\
{LADI-VTON}&0.049&0.928&&13.26&0.27&&\best{0.051}&0.922&&14.80&0.31&&0.089&0.868&&13.40&0.25\\
{\methodname} (\textbf{ours})
&\best{0.045}&\best{0.930}&&\best{11.45}&\best{0.16}&&\best{0.051}&\best{0.924}&&\best{13.27}&\textbf{0.24}&&\best{0.079}&\best{0.891}&&\best{12.61}&\best{0.19}\\
\shline
\end{tabular}
}
\caption{Comparison results on the Dress Code dataset. The best  results are highlighted in \best{bold}.
}
\label{tab:comp_dress}
\end{table*}

\begin{figure}[!ht]
    \centering
    \includegraphics[width=.85\linewidth]{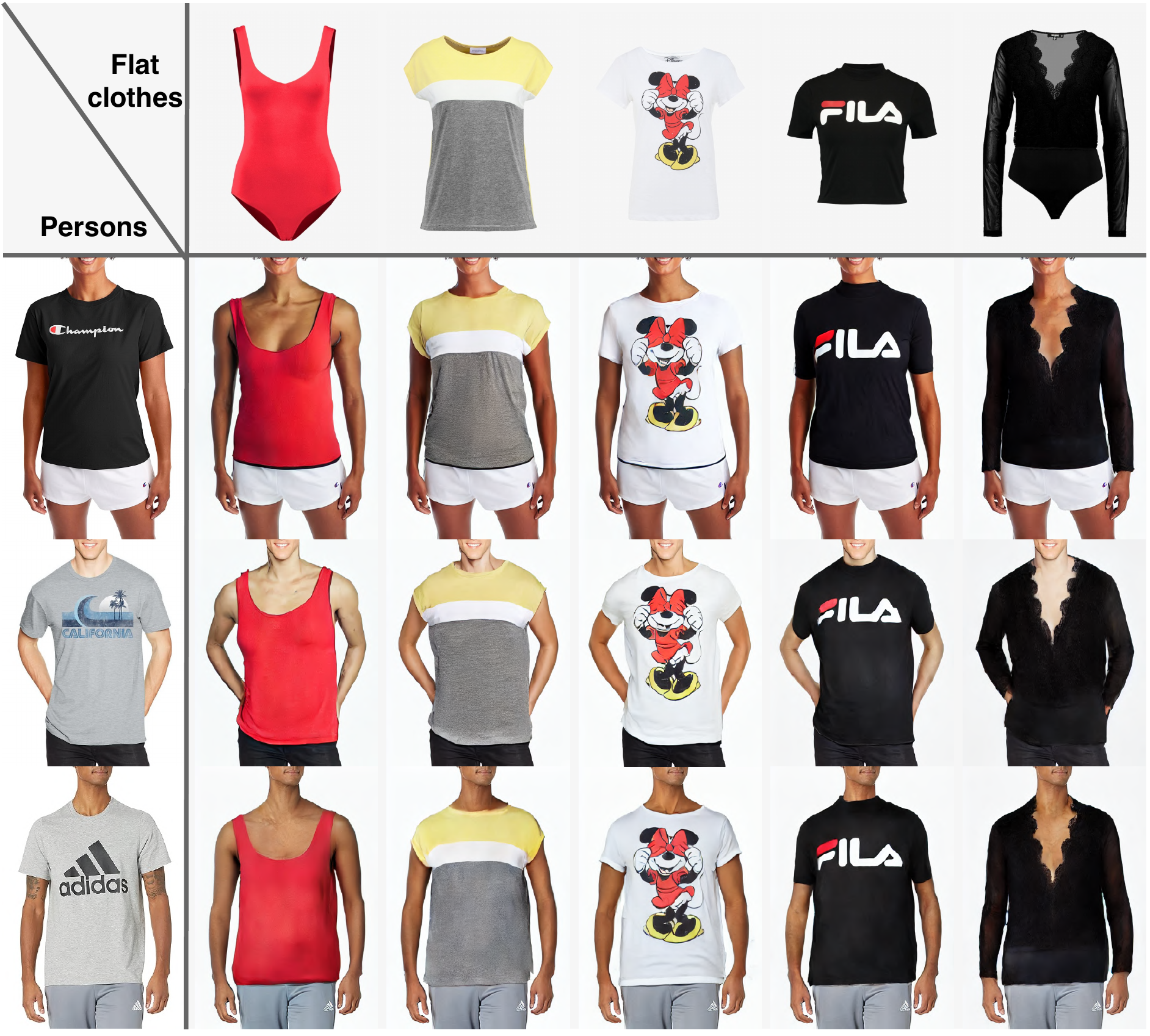}
    \caption{Qualitative results on real-world data.
    }
    \label{fig:real_data}
\end{figure}

\begin{figure}[!ht]
\centering

\includegraphics[width=1 \linewidth]{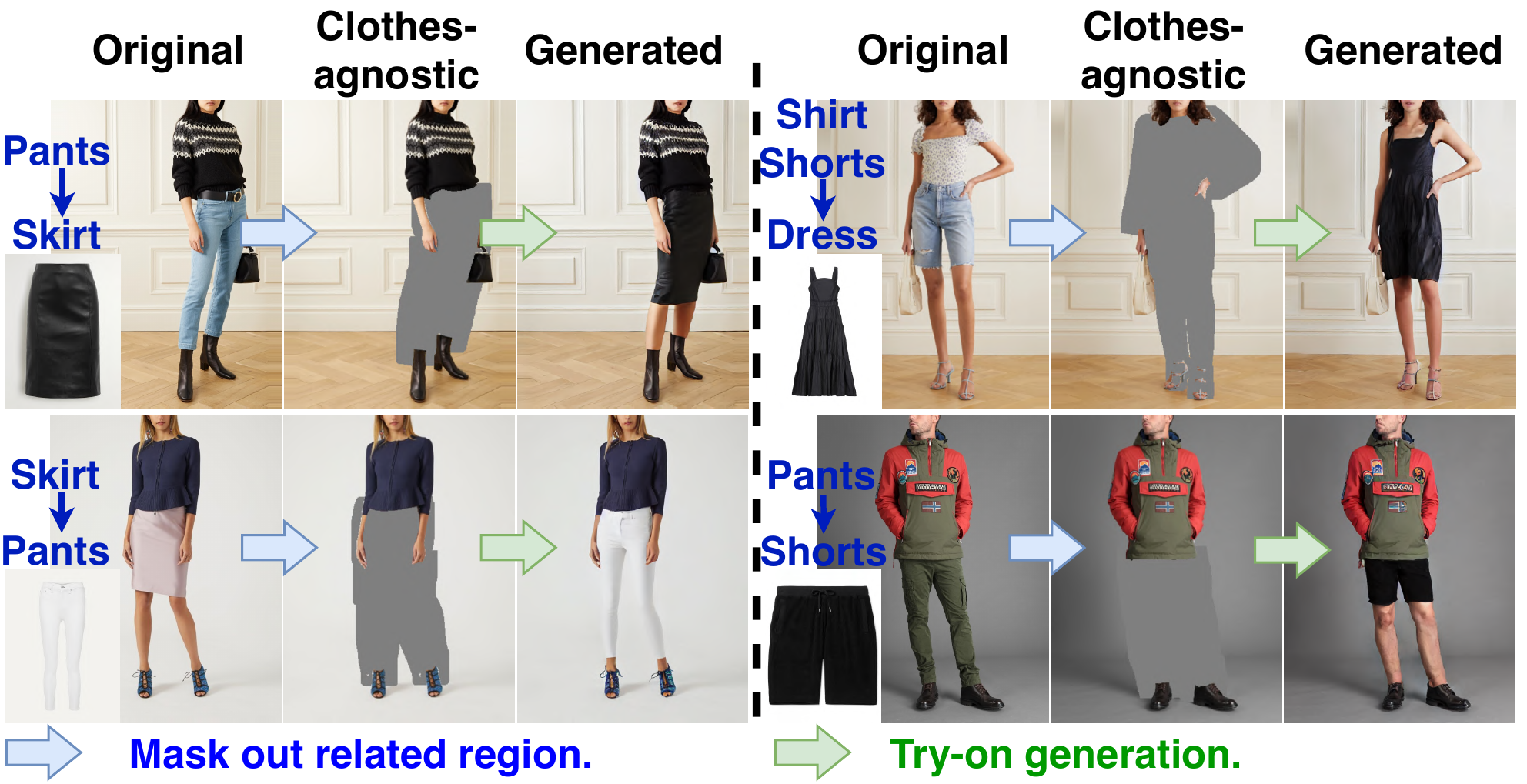}
\caption{Results of different source and target clothing types.}
\label{fig:similar_demo}
\end{figure}

\subsection{Comparison with SOTA Methods}
We comprehensively compare our \methodname  with several SOTA methods, which can be categorized into three groups: CNN-based, GAN-based, and diffusion-based. 
Specifically, the CNN-based methods are CP-VTON~\cite{wang2018toward} and CP-VTON+~\cite{minar2020cp}. 
For the GAN-based methods, we include VITON-HD~\cite{choi2021viton}, HR-VITON~\cite{lee2022high}, and GP-VTON~\cite{xie2023gp}. 
Moreover, we compare our \methodname with recently published SOTA diffusion-based methods: PAINT-BY-EXAMPLE~\cite{yang2023paint}, LADI-VTON~\cite{morelli2023ladi}, and DCI-VTON~\cite{gou2023taming}. Due to the extensive size of the Dress Code dataset, we only compare our \methodname with several SOTA methods, including GP-VTON, PAINT-BY-EXAMPLE, and LADI-VTON.

\paragraph{Quantitative comparison.} Tables~\ref{tab:comp_vton} and~\ref{tab:comp_dress} present quantitative results on the VITON-HD and Dress Code datasets, respectively, which show that our \methodname outperforms most competitors across various performance metrics under paired and unpaired  settings. We observe that diffusion-based methods perform better in terms of FID and KID, two realistic metrics. However, our \methodname not only maintains comparable realistic performance but also excels in faithfulness under the paired setting, \ie~LPIPS and SSIM.

\paragraph{Qualitative comparison.} Figures~\ref{fig:first},~\ref{fig:comp_vton}, and~\ref{fig:comp_dress}  present qualitative results, which demonstrate the superior performance of our \methodname in generating realistic try-on images while preserving faithful clothing details to the original flat clothes. 
Although CNN-based methods generate try-on images that more closely resemble the original flat clothes, they lack realism and don't achieve a photo-like quality. 
With adversarial training to enhance the initially warped clothes, GP-VTON stands out with superior performance. However, its qualitative results suggest a relatively simple composition of warped clothes and persons due to GAN's intrinsic model collapse problem. 
Among diffusion-based methods, PAINT-BY-EXAMPLE and LADI-VTON generate more realistic images. However, the stochastic nature of conventional diffusion models poses a challenge in faithfully preserving crucial clothing details, often leading to substantial distortions. DCI-VTON faces challenges in correctly adapting the clothes' style to the person, though generating relatively faithful images. 
In contrast, our \methodname is able to generate more realistic images while effectively handling complex style, pattern, and text; please refer to Appendix~\ref{app:qua} for more qualitative results.

\begin{figure*}[!ht]
    \centering
    \includegraphics[width=.95\linewidth]{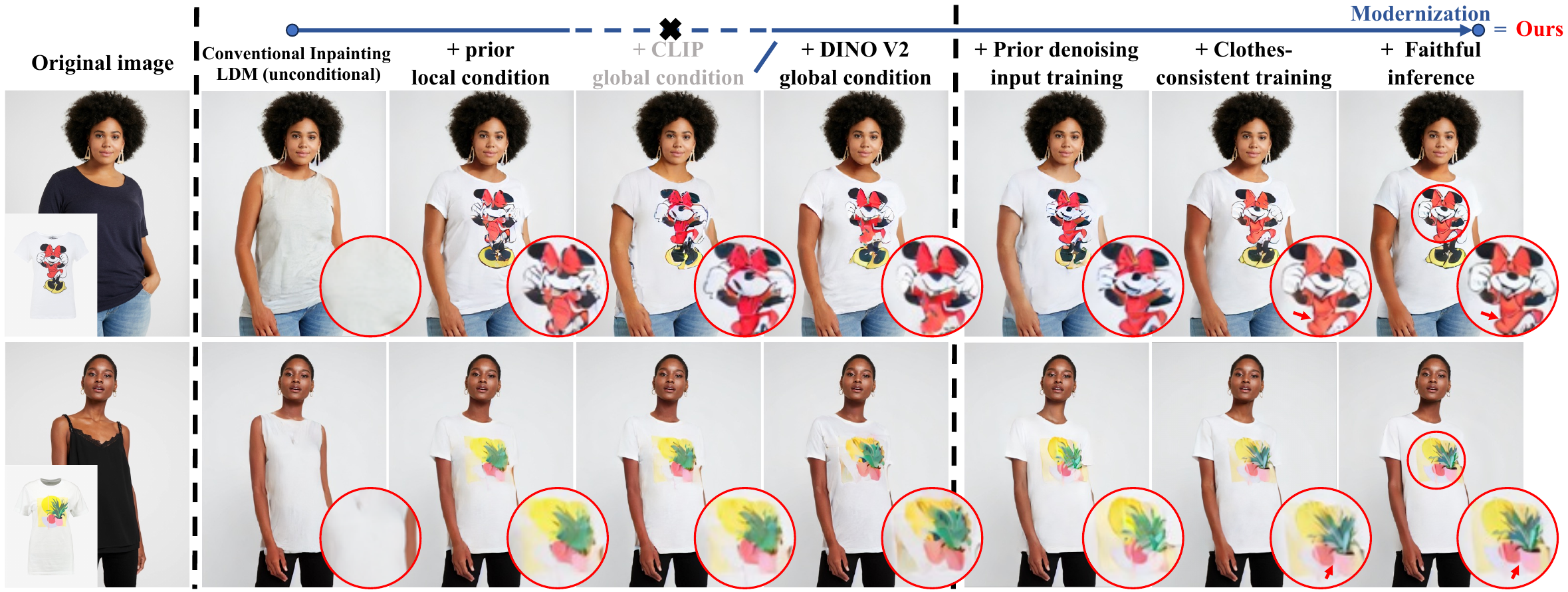}
    \caption{Ablation qualitative results for modernizing our \methodname on the VITON-HD dataset. Best viewed when zoomed in.}
    \label{fig:ablation_vton}
\end{figure*}

\paragraph{Application to real-world data.} \figref{fig:real_data} also presents qualitative results on real-world data from the Amazon website, illustrating the robustness of our \methodname. Our method effectively generates corresponding try-on images for different persons with diverse flat clothes, ensuring photo-realistic quality with faithful clothing details.

\paragraph{More challenging try-on situations.} We further examine our \methodname on challenging try-on situations, such as different types of source and target clothes, by simply \emph{masking out the region related to the source and target clothes}. The challenging try-on results in Figure~\ref{fig:similar_demo} suggest the strong effectiveness of our \methodname.

\subsection{Ablation Study}

Here, we conduct detailed ablation studies to show the effectiveness of different conditions and proposed components for modernizing our \methodname.
\figref{fig:ablation_vton} and \tabref{tab:abla_vton} present ablation qualitative and quantitative results, respectively. 

\begin{table}[!t]
\centering
\resizebox{\linewidth}{!}{
\begin{tabular}{lccccc}
\shline
\multirow{2}{*}{\textbf{Methods}} &\multicolumn{2}{c}{\textbf{Paired}} &&\multicolumn{2}{c}{\textbf{Unpaired}} \\
&\textbf{LPIPS}$\downarrow$&\textbf{SSIM}$\uparrow$ &&\textbf{FID}$\downarrow$ &\textbf{KID}$\downarrow$ \\
\cline{1-1}\cline{2-3}\cline{5-6}
Conventional inpainting LDM&0.158&0.788&&34.58&3.41\\
\: + Prior local condition (Sec.~\ref{sec:diff_main}) &0.127&0.864&&13.43&0.34\\
\: \: \textcolor{Gray!80}{+ CLIP global condition} &\textcolor{Gray!80}{0.121} &\textcolor{Gray!80}{0.870} &&\textcolor{Gray!80}{13.02}&\textcolor{Gray!80}{0.32}\\
\: \: + DINO V2 global condition&0.114&0.874&&12.24&0.29\\
\cline{1-1}\cline{2-3}\cline{5-6}
\: \: \: + Prior denoising input training (Sec.~\ref{sec:diff_main})&0.094&0.876&&10.60&0.27\\
\: \: \: \: + Clothes-consistent supervision (Sec.~\ref{sec:takeoff})&0.082&0.883&&\,\,8.90&0.15\\
\: \: \: \: \: + Faithful inference (Sec.~\ref{sec:overview}) &\best{0.080}&\best{0.886}&&\,\,\best{8.81}&\best{0.13}\\
\shline
\end{tabular}
}
\caption{Ablation quantitative results on the VITON-HD dataset. The best  results are highlighted in \best{bold}.}
\label{tab:abla_vton}
\end{table}

\paragraph{Ablation on different conditions.} We take the conventional unconditional inpainting LDM as our initial baseline. First, we enhance it with the prior local condition $\widehat{\img{T}}^w$. We observe that the generated try-on image can capture most coarse-grained clothing details due to the prior information provided by warped clothes. However, the style of complex patterns is significantly distorted. Second, by integrating the CLIP or DINO V2 global conditions, there is a notable improvement in the style of these complex patterns. Notably, DINO V2 outperforms CLIP, which can be attributed to its more discriminative global feature extraction.

\paragraph{Ablation on proposed components.} Here, we take the inpainting LDM with the prior local condition and DINO V2 global condition as the initial baseline. We first investigate the impact of the additional training for the prior denoising input. By including this training process, we observe that the style of complex patterns is largely preserved. However, there still remains a challenge in preserving fine clothing details.

\begin{figure}[t]
\centering
\includegraphics[width=.89\linewidth]{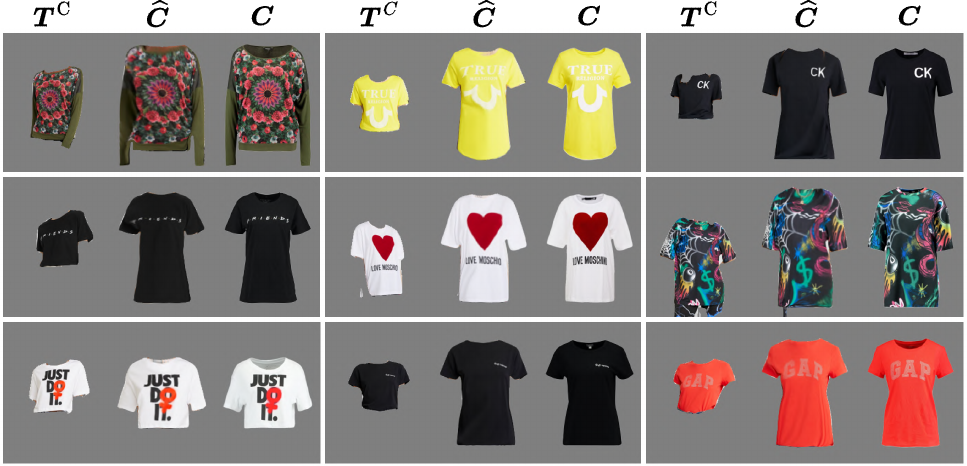}
 \caption{Clothes-consistent results. $\img{T}^C$, $\widehat{\img{C}}$, and $\img{C}$ denote the real clothes-parsed image,  estimated flat clothes image, and  real flat clothes image, respectively. Best viewed when zoomed in.}
\label{fig:takeoff_result} 
\end{figure}

Second, we show the effect of clothes-consistent supervision in \figref{fig:takeoff_result}, including the real clothes-parsed image $\img{T}^C$, estimated flat clothes image $\widehat{\img{C}}$, and real flat clothes image $\img{C}$. We find that our clothes flattening network effectively transforms the clothes from the try-on state to the flat state. By comparing the estimated and the original flat clothes, we observe that the original flat clothes have more fine clothing details, which further validates the importance of providing clothes-consistent supervision to guide the try-on diffusion process towards higher faithfulness to the original flat clothes. In summary, both the quantitative and qualitative results  demonstrate that introducing the clothes flattening network into the training improves the capability of the try-on diffusion UNet to preserve more faithful clothing details.

In addition, we observe that while the faithful inference contributes not as much as other components to the quantitative results in \tabref{tab:abla_vton}, it significantly enhances the faithfulness of generated clothing details, as shown in \figref{fig:ablation_vton}, without introducing additional computational burdens.  Please refer to Appendix~\ref{app:abla} for more ablation results about FreeU, and 
Appendix~\ref{app:discuss} for more discussions, including the acceleration advantage, the limitation, and the social impact. 

%% file: sec/5_con.tex
\section{Conclusion}

In this paper, we proposed a novel faithful latent diffusion model for virtual try-on. With the introduced faithful clothes priors and clothes-consistent faithful supervision, the proposed \methodname can significantly alleviate the unfaithful generation issue caused by the diffusion stochastic nature and latent supervision in LDM. 
In addition, the devised clothes-posterior sampling for faithful inference can further improve the model performance. 
Extensive experimental results on two popular VTON benchmarks validate the superior performance of our \methodname---generating photo-realistic try-on images with faithful clothing details.

%% file: sec/6_supp.tex
\section*{Appendix}
\setcounter{figure}{0}
\setcounter{section}{0}
\setcounter{table}{0}
\renewcommand{\thefigure}{A\arabic{figure}}
\renewcommand{\theequation}{A\arabic{equation}}
\renewcommand{\thesection}{ \Roman{section}}
\renewcommand{\thetable}{A\arabic{table}}

\section{More Implementation Details}
\label{app:imple}
This section presents visualizations and detailed pipelines of $\img{P}^\text{a}$ and $\widehat{\img{T}}^\text{w}$, the clothes warping procedure, and more training and sampling details.

\begin{figure}[h]
\centering

\includegraphics[width=1\linewidth]{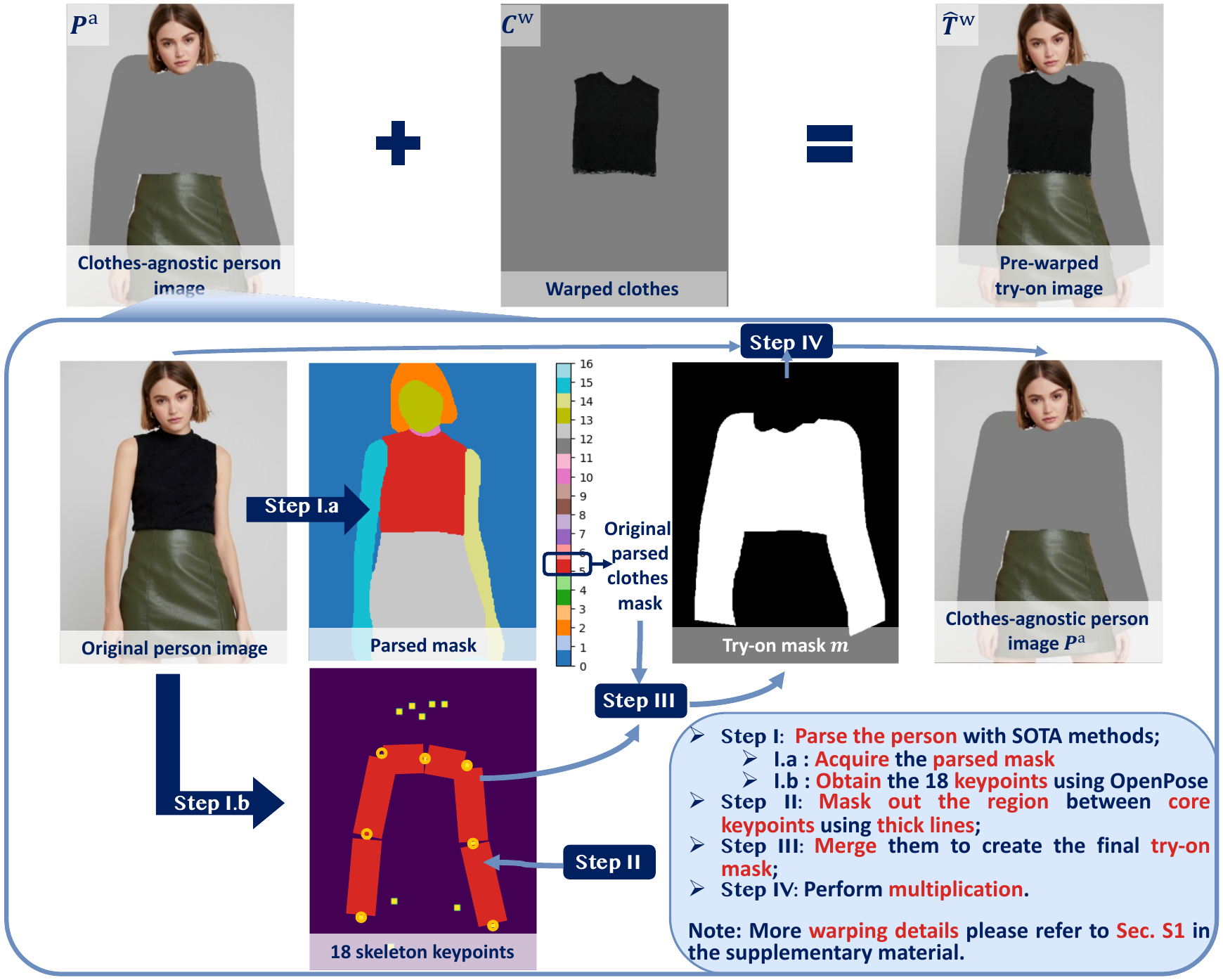}
\caption{Visualizations and detailed pipelines of $\img{P}^\text{a}$ and $\widehat{\img{T}}^\text{w}$.}
\label{fig:pa_pipeline}
\end{figure}

\paragraph{Visualizations and detailed pipelines of $\img{P}^\text{a}$ and $\widehat{\img{T}}^\text{w}$.} The clothes-agnostic person image $\img{P}^\text{a}$ refers to the person image with the try-on region being masked out. The pre-warped try-on image $\widehat{\img{T}}^\text{w}$ 
is simply obtained by adding the warped clothes image $\img{C}^\text{w}$ to $\img{P}^\text{a}$, \ie~$\widehat{\img{T}}^\text{w}=\img{C}^\text{w}+\img{P}^\text{a}$. \figref{fig:pa_pipeline} presents the visualizations and detailed 
processing pipelines of these two definitions to facilitate the understanding. Of note, all preprocessing pipelines of our \methodname follow previous VTON research~\cite{lee2022high}. We will certainly make them clearer in the future version.

\begin{figure}[!h]
    \centering
    \includegraphics[width=1 \linewidth]{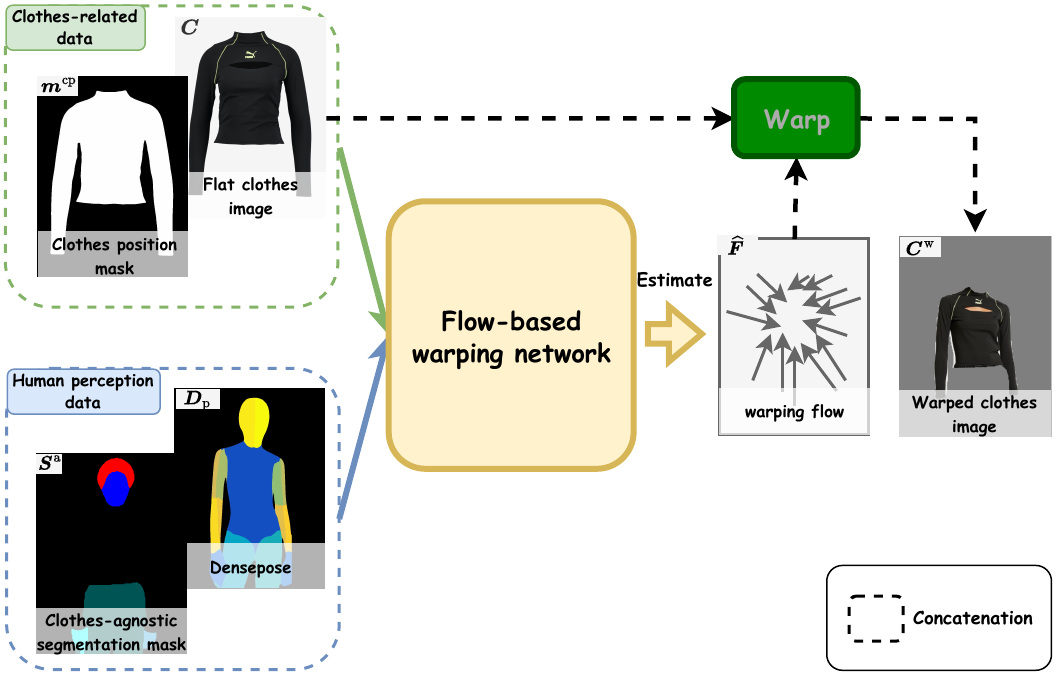}
    \caption{Workflow of the clothes warping procedure.}
    \label{fig:warp}
\end{figure}

\paragraph{Clothes warping procedure.}
\figref{fig:warp} presents the detailed clothes warping procedure. 
To adapt the flat clothes image $\img{C}$ to fit a  person $\img{P}$, we exploit the appearance flow-based network proposed in~\cite{gou2023taming,ge2021parser} to estimate the warping flow $\widehat{\img{F}}$, which is then utilized to warp the original flat clothes $\img{C}$, generating the corresponding warped clothes image $\img{C}^\text{w}$.

Specifically, our appearance flow-based warping network processes both \emph{human perception data} and \emph{clothes-related data} to estimate the warping flow $\widehat{\img{F}}$. The human perception data comprises the clothes-agnostic segmentation mask $\img{S}^\text{a}$ and dense pose image $\img{D}_\text{p}$. Meanwhile, the clothes-related data include the flat clothes image $\img{C}$ and flat clothes position mask $\img{m}^\text{cp}$. Then, the warping network takes them as input to estimate the final warping flow $\widehat{\img{F}}$, generating the corresponding warped clothes image $\img{C}^\text{w}$.

Following the training procedure of the proposed clothes flattening  network and other appearance flow-based methods~\cite{ge2021parser}, we optimize the warping network with the following loss function, including both pixel- and flow-level components:
\begin{align}\loss{Warp}=\loss{1}^\text{w}+\lamda{per}^\text{w}\loss{per}^\text{w}+\lamda{sec}^\text{w}\loss{sec}^\text{w}+\lamda{TV}^\text{w}\loss{TV}^\text{w},
\end{align}
where $\lamda{*}^\text{w}$ are hyperparameters to adjust the weights among different loss components. We set $\lamda{per}^\text{w}=0.2$, $\lamda{sec}^\text{w}=0.01$, and $\lamda{TV}^\text{w}=6.0$, in line with the settings suggested by~\cite{gou2023taming,ge2021parser}.

\paragraph{Training and sampling details.}
In our experiments, we first train the clothes flattening network and then freeze its weights to train our try-on diffusion UNet.

First, we train the clothes flattening network with a mini-batch size of 32 over 100 epochs. Specifically, we start with an initial learning rate of $5.0\times{10}^{-5}$ for the first 30 epochs, which is then linearly decreased for the remaining epochs. The hyperparameters $\lamda{per}$, $\lamda{sec}$, and $\lamda{TV}$ in Eq.~(4) are set as 0.1, 10, and 0.01, respectively, ensuring consistency in the order of magnitude. Note that both the warping and clothes flattening networks are trained with a  resolution of size $256\times 192$, aligning with existing appearance flow-based methods~\cite{bai2022single,ge2021parser,he2022style}. When needed, we upsample the predicted warping or flattening flow to the corresponding size.

Subsequently, we train the try-on diffusion UNet over $200,000$ iterations, with a mini-batch size of $8$. The initial learning rate is  $1.0\times {10}^{-6}$, which is gradually increased to $2\times{10}^{-5}$ over the first $3000$ iterations. The hyperparameter $\lamda{cons}$ is empirically set as $0.15$, also ensuring consistency in order of magnitude. The try-on diffusion UNet is initialized with the official pre-trained weights of Paint-by-example~\cite{yang2023paint}, offering a basic inpainting ability. In line with the standard SD framework, classifier-free guidance~\cite{ho2022classifier} with an unconditional probability of $0.2$ is used for the global condition.

During the sampling phase, we employ the DPM solver~\cite{lu2022dpm} method with 50 steps. Following the official adjusted suggestion of FreeU~\cite{si2023freeu}, we empirically adjust the weights of the backbone ($b$) and skip-connection ($s$) features at the first two stages with the value of $b_1=1.1$, $b_2=1.2$, $s_1=0.9$, $s_2=0.6$; please refer to Appendix~\ref{app:abla} for more FreeU ablation results. In addition, for the implementation of competing methods presented in this paper, we use pre-trained weights if available; otherwise, we train the models following their official codes.

\section{More Qualitative  Results}
\label{app:qua}
To further demonstrate the effectiveness of our \methodname, this section presents more qualitative results on the VITON-HD and Dress Code datasets.

\begin{figure*}[!ht]
    \centering
    \includegraphics[width=1 \linewidth]{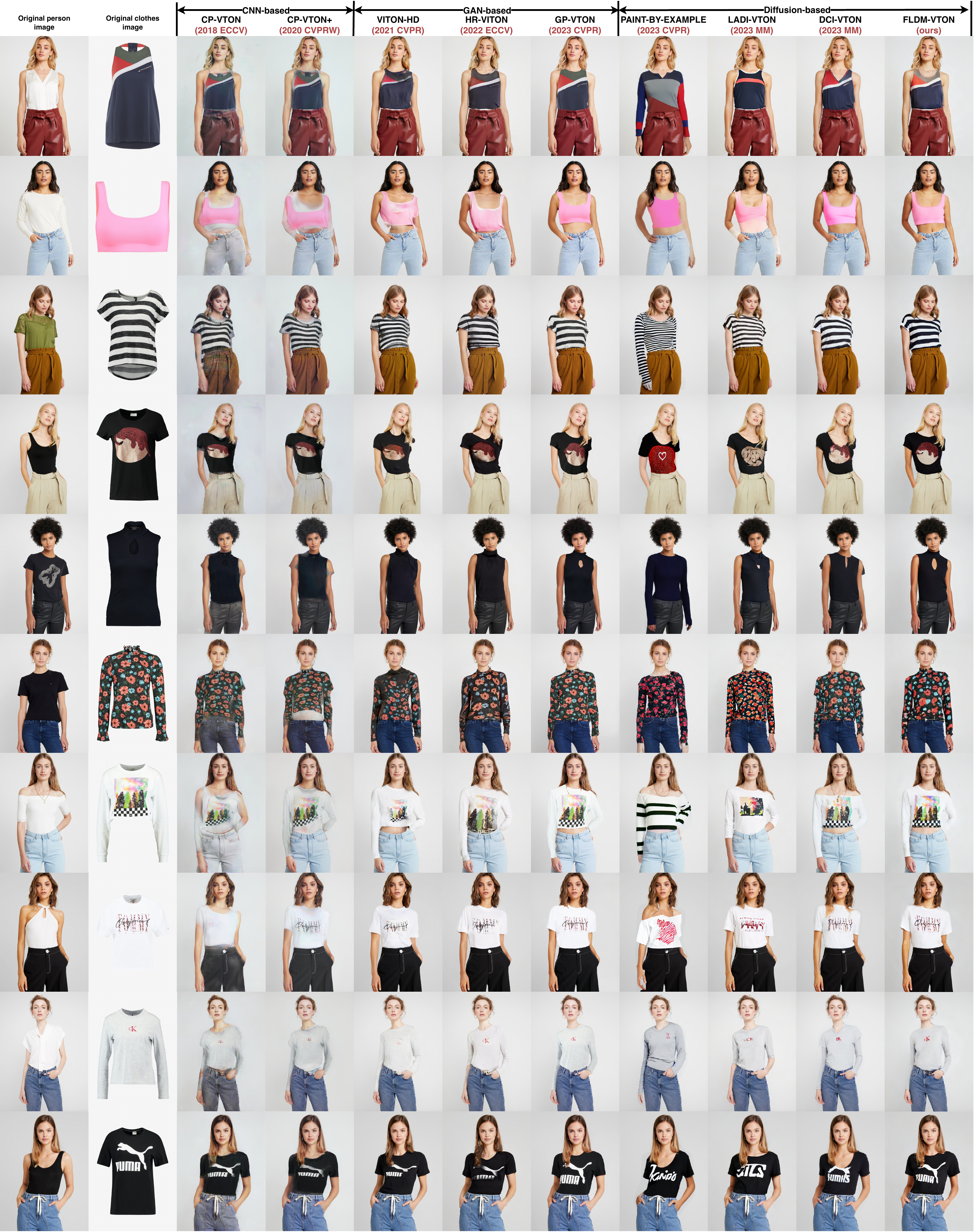}
    \caption{More qualitative results of different methods and our \methodname on the VITON-HD dataset.}
    \label{fig:supp_viton}
\end{figure*}

\paragraph{Results on the VITON-HD dataset.}
\figref{fig:supp_viton} presents more qualitative results on the VITON-HD dataset. Specifically, we comprehensively compare our \methodname with 8 SOTA competing methods, including CP-VTON~\cite{wang2018toward}, CP-VTON+~\cite{minar2020cp}, VITON-HD~\cite{choi2021viton}, HR-VITON~\cite{lee2022high}, GP-VTON~\cite{xie2023gp}, Paint-by-example~\cite{yang2023paint}, Ladi-VTON~\cite{morelli2023ladi}, and DCI-VTON~\cite{gou2023taming}.

\begin{figure*}[!ht]
    \centering
    \includegraphics[width=0.9 \linewidth]{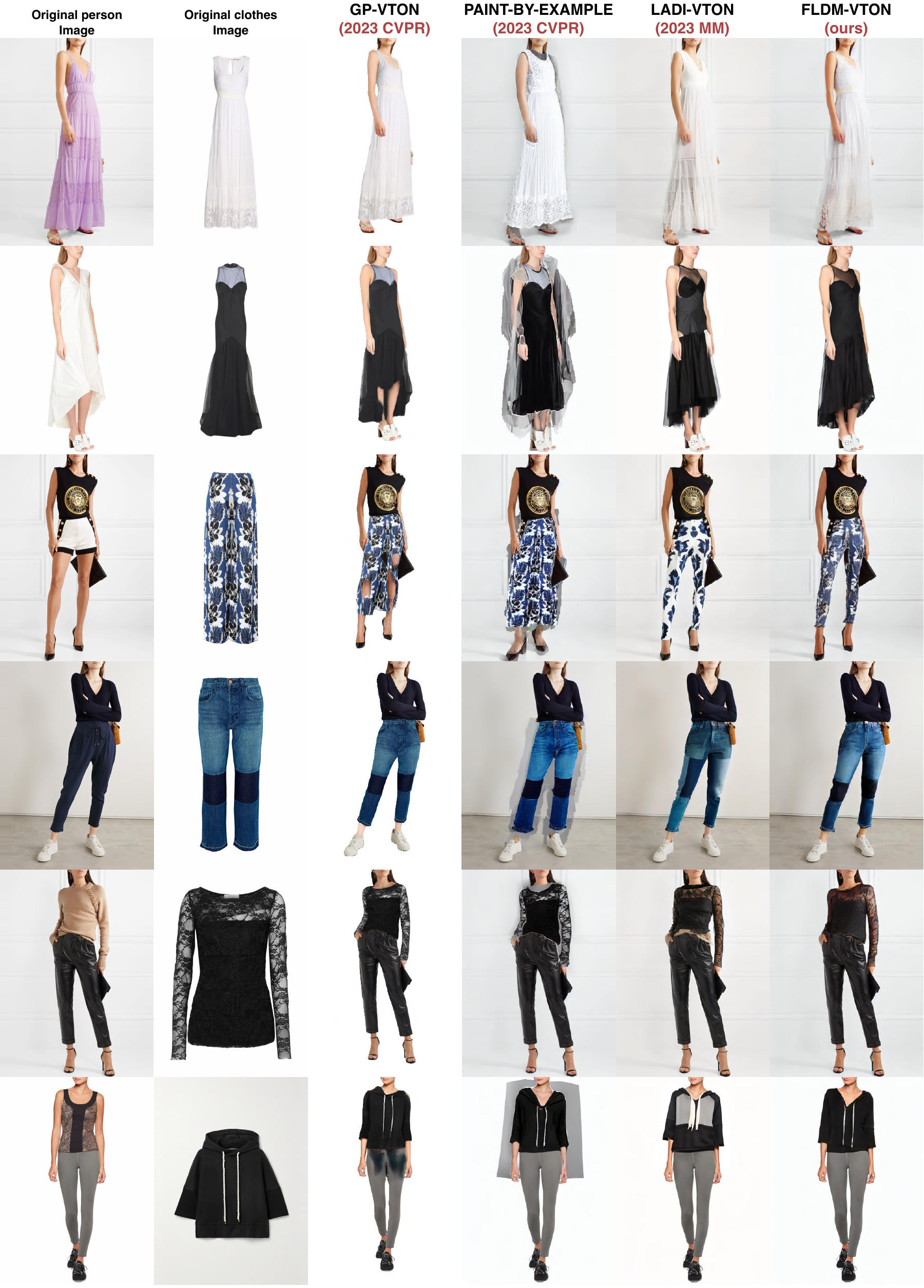}
    \caption{More qualitative results of different methods and our \methodname on the Dress Code dataset.}
    \label{fig:supp_Dress}
\end{figure*}

\paragraph{Results on the Dress Code dataset.}
\figref{fig:supp_Dress} presents more qualitative results of our \methodname and other competing methods on the Dress Code dataset.  Due to the extensive size of the Dress Code dataset, we only compare our FLDM-VTON with several SOTA methods that have disclosed pre-training weights, including GP-VTON, Paint-by-example, and Ladi-VTON.

\section{More Ablation Study}
\label{app:abla}
\begin{figure}[!h]
    \centering
    \includegraphics[width=1\linewidth]{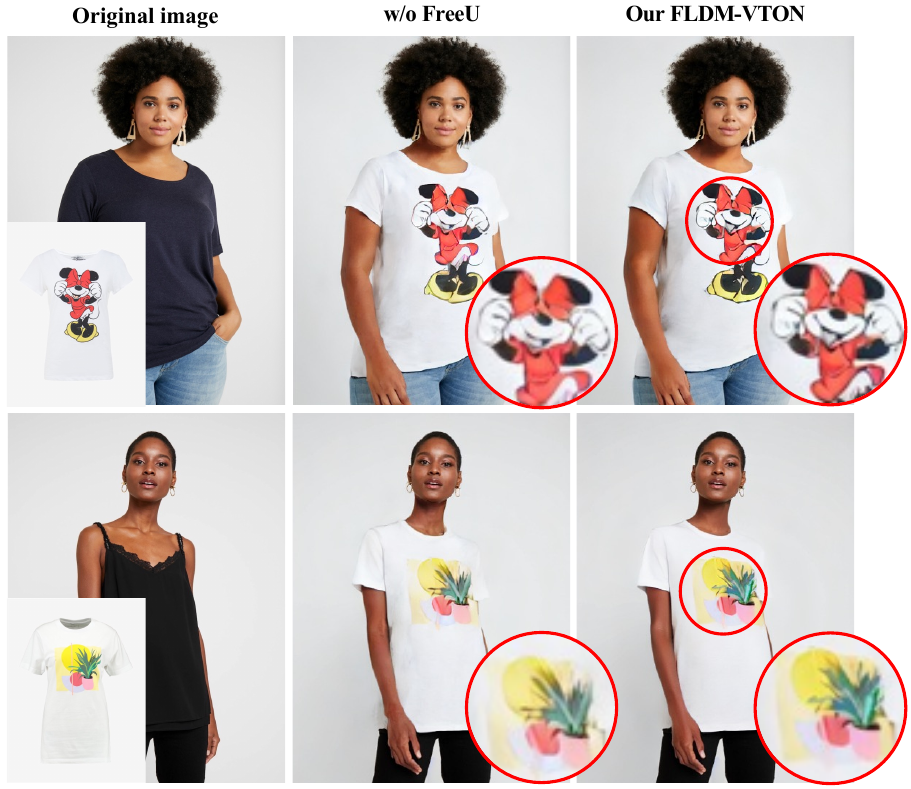}
    \caption{Qualitative  ablation  results of FreeU used in our \methodname.}
    \label{fig:freeu}
\end{figure}

\figref{fig:freeu} 
 and \tabref{tab:abla_vton_sup}  present the qualitative and quantitative   ablation results on FreeU, respectively. We observe that FreeU does not contribute as much as other proposed components. However, 
the experimental results show that the use of FreeU is able to enhance the sample quality of diffusion models without introducing any additional computational costs, as demonstrated in the original paper~\cite{si2023freeu}.

\begin{table}[!h]
\centering
\resizebox{\linewidth}{!}{
\begin{tabular}{lccccc}
\shline
\multirow{2}{*}{\textbf{Methods}} &\multicolumn{2}{c}{\textbf{Paired}} &&\multicolumn{2}{c}{\textbf{Unpaired}} \\
&\textbf{LPIPS}$\downarrow$&\textbf{SSIM}$\uparrow$ &&\textbf{FID}$\downarrow$ &\textbf{KID}$\downarrow$ \\
\cline{1-1}\cline{2-3}\cline{5-6}
Our \methodname &\best{0.080}&\best{0.886}&&\,\,\best{8.81}&\best{0.13}\\
\quad w/o FreeU&0.081&0.884&&\,\,8.83&\best{0.13}\\
\shline
\end{tabular}
}
\caption{Quantitative ablation results of FreeU used in our \methodname. The best results are highlighted in \best{bold}.}
\label{tab:abla_vton_sup}
\end{table}
\section{Discussions}
\label{app:discuss}
\begin{figure*}[t]
    \centering
    \includegraphics[width=1 \linewidth]{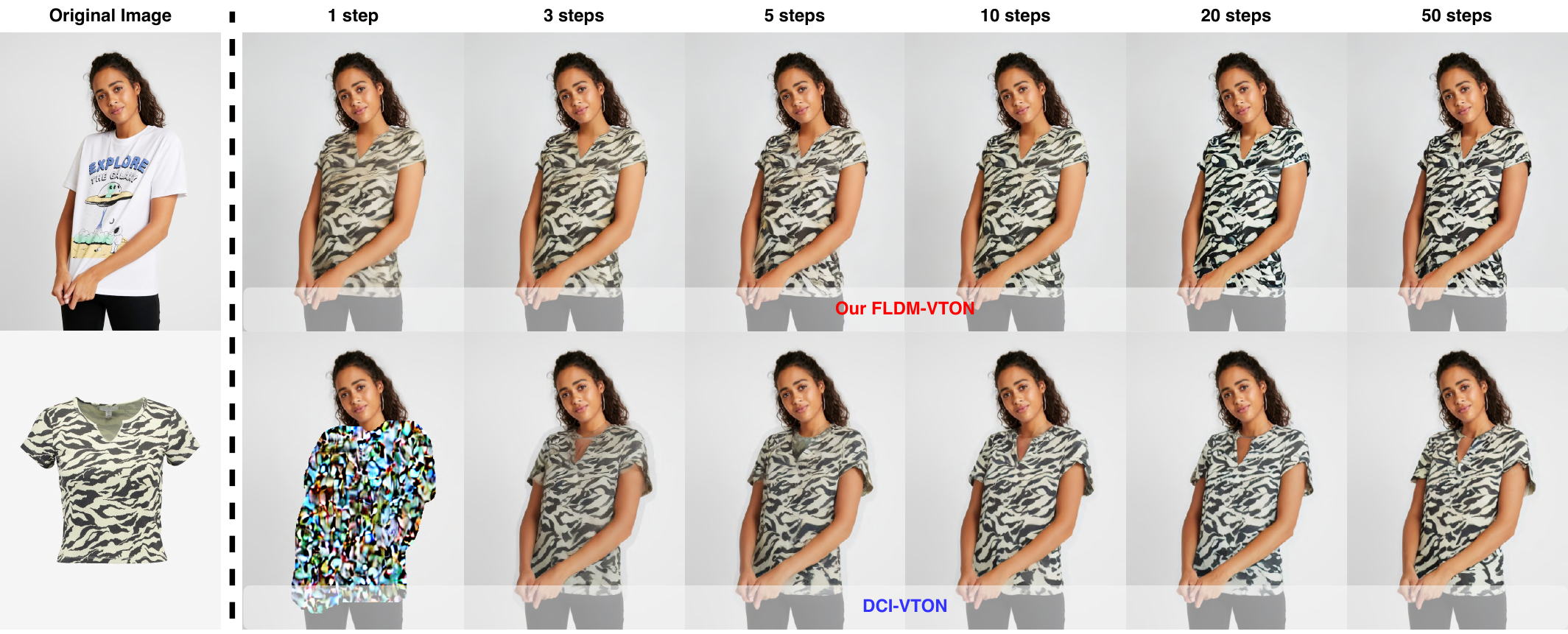}
    \caption{Comparison results of DCI-VTON and our \methodname using the same DDIM sampler  with various sampling steps.}
    \label{fig:accelerated}
\end{figure*}

\paragraph{Acceleration advantage.} We employ the same DDIM sampler~\cite{song2020denoising}  with various sampling steps for inference in both DCI-VTON and our \methodname, as illustrated in~\figref{fig:accelerated}. As the number of sampling steps increases, we observe that DCI-VTON exhibits an \emph{unstable and inconsistent} sampling process. Notably, the open neckline of the presented person intermittently appears and disappears. In contrast, our \methodname can generate photo-realistic try-on images while preserving most clothing details with fewer sampling steps. Moreover, our sampling process exhibits a more stable and consistent refinement phenomenon.

\begin{figure*}[t]
    \centering
    \includegraphics[width=0.6 \linewidth]{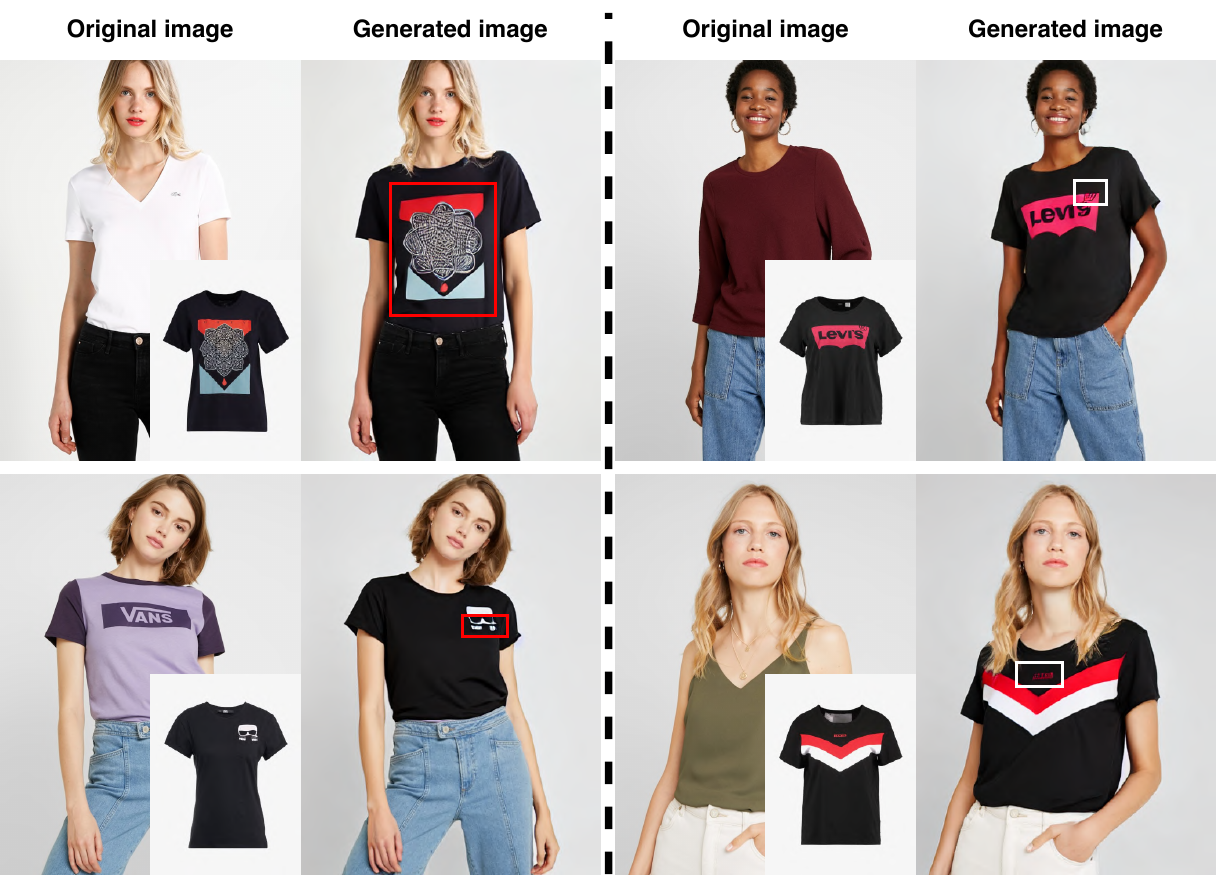}
    \caption{Some failure cases by our \methodname.}
    \label{fig:limi}
\end{figure*}

\paragraph{Limitation.} While our \methodname exhibits promising performance, we acknowledge some limitations in preserving \emph{extremely small or complex} logos and patterns. \figref{fig:limi} presents some failure cases. Despite its ability to generate high-quality try-on images and preserve most clothing details, handling intricate scenarios remains a struggle. This limitation arises from the inherent loss of vital clothing information during the latent diffusion process, which may be alleviated by performing the diffusion process in pixel space with sufficient computational resources or providing a more robust pretrained LDM.

\paragraph{Social impact.} The VTON technology has a significant social impact by revolutionizing the retail experience. It offers consumers a personalized shopping journey, reducing clothing returns and waste. In addition, it supports sustainable practices in the fashion industry by minimizing the need for physical fittings. Moreover, it fosters social engagement as users can share their virtual try-on experiences on social platforms, creating a dynamic and interactive online community.

